\documentclass[a4paper]{ifacconf}

\usepackage{graphicx}
\usepackage{natbib} 
\usepackage{amsmath}
\usepackage{amssymb}
\usepackage{bm}
\usepackage{tikz}
\usepackage[hyphens]{url}

\let\scalarEta\eta
\let\scalarNu\nu
\let\scalarTau\tau
\let\scalarOmega\omega

\renewcommand{\eta}{\bm{\scalarEta}}
\newcommand{\etaDot}{\bm{\dot{\scalarEta}}}
\renewcommand{\nu}{\bm{\scalarNu}}
\newcommand{\nuDot}{\bm{\dot{\scalarNu}}}
\renewcommand{\tau}{\bm{\scalarTau}}
\renewcommand{\omega}{\bm{\scalarOmega}}
\newcommand{\bv}{\bm{v}}
\newcommand{\vp}{\bm{v}_p}

\newcommand{\J}{\bm{J}}
\newcommand{\M}{\bm{M}}
\newcommand{\C}{\bm{C}}
\newcommand{\D}{\bm{D}}
\newcommand{\g}{\bm{g}}
\newcommand{\R}{\mathbb{R}}
\renewcommand{\S}{\bm{S}}

\newcommand{\figref}{Figure~\ref}

\begin{document}
\begin{frontmatter}

\title{Combining Moving Mass Actuators and Manoeuvring Models for Underwater Vehicles: A Lagrangian Approach}

\author[1]{Alexander B. Rambech} 
\author[1]{Ivar B. Saksvik}
\author[1,2]{Vahid Hassani}

\address[1]{Oceanlab, Oslo Metropolitan University (OsloMet),
   Oslo, Norway (e-mail: alexander.rambech@oslomet.no, ivar.saksvik@oslomet.no, vahid.hassani@oslomet.no).}
\address[2]{Department of Ships and Ocean Structures, SINTEF Ocean, Trondheim, Norway}

\begin{abstract}
In this paper, we present a Newton-Euler formulation of the equations of motion for underwater vehicles with an interntal moving mass actuator. 
Furthermore, the moving mass dynamics are expressed as an extension to the manoeuvring model for underwater vehicles, originally introduced by \cite{Fos:91}. 
The influence of the moving mass is described in body-frame and included as states in both an additional kinematic equation and as part of the coupled rigid-body kinetics of the underwater vehicle. 
The Coriolis-centripetal effects are derived from Kirchhoff's equations and the hydrostatics are derived using first principals.
The proposed Newton-Euler model is validated through simulation and compared with the traditional Hamiltonian internal moving mass actuator formulation.
\end{abstract}

\begin{keyword}
Underwater Vehicles, Nonlinear Models, Manoeuvring Models, Moving Mass Actuators
\end{keyword}

\end{frontmatter}

\section{Introduction}
Internal moving mass actuators enable the control of a vehicle's centre of gravity (CG). 
For a craft where control surfaces have little or no effect on attitude, moving mass actuators can be exploited to increase articulation. 
Examples of such systems are spacecraft in orbit and slow-moving underwater vehicles like buoyancy propelled gliders. 
Moreover, for underwater vehicles, moving mass actuators enable flight without inducing additional drag from control surfaces. 
Moving mass actuators can therefore be used to increase endurance of autonomous underwater vehicles (AUVs).
This has been demonstrated by \cite{Hob:12} and \cite{Coz:19}, among others.

From classical and fluid mechanics, there are several ways to derive a model of an underwater vehicle with an internal moving mass.
\cite{Leo:01} used Hamiltonian mechanics in order to derive the kinetics of the ROUGE glider with forces on the moving mass as control inputs. 
This formulation was further generalized by \cite{Woo:02}. 

These models are elegant solutions to the modelling of underwater vehicles with internal moving mass actuators and provide valuable insight into their dynamics and formulations where control inputs are expressed as direct forces on the moving mass subsystem. 

A popular modelling framework for marine craft is the vectorial Newton-Euler equations of motion introduced by \cite{Fos:91}. 
The clean notation and strong ties to first principals, makes the equations an attractive choice for modelling of underwater vehicles. 
Furthermore, the matrix formulation contains properties that are well suited for simulation and control design.

Moving masses have been used in tandem with these vectorial models in several ways. 
In \cite{Sak:23} the movement of the mass is not present in the kinetic equation, instead the position of the moving mass is treated as a control input.
With this approach there is no need to alter the kinetics of the model by including the coupled relationship between the moving mass and AUV. 
However, including the actuator in such a manner does not capture the coupled dynamics between the moving mass and the rest of the system. 

In \cite{Fos:21} the equations found in \cite{Woo:02} are put in the context of the vectorial equations of motion, but the proper link to Newton-Euler is never established.

Presented in this article is a Newton-Euler formulation of the equations of motion of underwater vehicles with moving mass actuators, on a vectorial form. 
The model is derived using Lagrangian mechanics and verified through a simulation example of the Remus 100 AUV. We also compare the simulation results of our model with simulations of the Hamiltonian formulation found in \cite{Woo:02}.

\section{Manoeuvring model}
\label{sec:cams25-modelling}
\subsection{Kinematics}
\label{sec:cams25-kinematics}
\begin{figure}[t]
    \centering
    \includegraphics[width=0.4\textwidth]{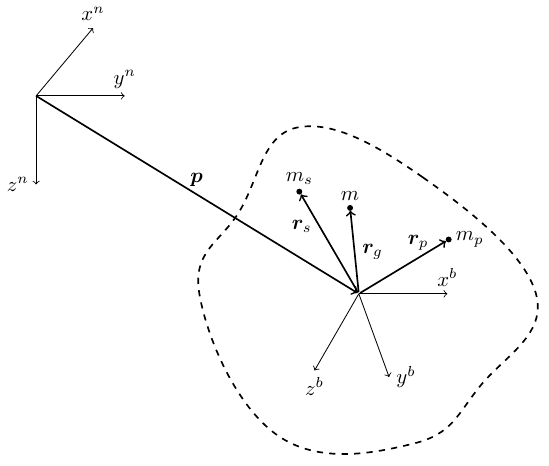}
    \caption{Kinematics of marine craft with moving mass in relation to earth-fixed inertial frame.}
    \label{fig:cams25-kinematics}
\end{figure}

Following the notational convention used by \cite{Fos:21}, the vessel kinematics with respect to the earth-fixed inertial frame is given by:
\begin{equation*}
    \etaDot = \J(\eta)\nu
\end{equation*}
where $\etaDot$ denotes the derivative of the generalized vehicle coordinates given in the earth-fixed inertial frame, $\eta = [\bm{p}^\top, \bm{\Theta}^\top] = \left[ x, y, z, \phi, \theta, \psi \right]^\top \in \R^3 \times \mathbb{T}^3$, the vehicle velocities given in the vehicle-fixed frame are denoted $\nu = [ \bv^\top, \omega^\top ]^\top = [ u, v, w, p, q, r ]^\top \in \R^6$ and $\J(\eta) = \text{diag}\left\{ \bm{R}(\bm{\Theta}), \bm{T}(\bm{\Theta}) \right\}$ is the transformation relating velocities in the two reference frames. 
Here, $\bm{R}: \mathbb{T}^3 \rightarrow \text{SO}(3)$ and $\bm{T}: \mathbb{T}^3 \rightarrow \R^{3 \times 3}$ are the linear and angular velocity transforms between the vehicle-fixed and the inertial frame, respectively. \figref{fig:cams25-kinematics} illustrates the relationship between the two reference frames.

The vehicle-fixed frame, also referred to as body-frame, has its origin at the centre of orientation (CO) of the vehicle which is assumed fixed and located in the middle of the vessel. 
The centre of buoyancy (CB) is located at the CO and assumed fixed as well.

The moving mass actuator, with point mass $m_p$, is located at $\bm{r}_p = [ x_p, y_p, z_p]^\top \in \R^3$ compared to vehicle origin. 
The moving mass also impacts the location of the total point mass of the vehicle $m$.
This is located at:
\begin{equation*}
    \bm{r}_g = \frac{m_{s}\bm{r}_{s} + m_p\bm{r}_p}{m_{s} + m_p}
\end{equation*}
where $m_{s}$ denotes the mass of the remaining rigid body of the vessel with stationary position $\bm{r}_{s} \in \mathbb{R}^3$.
The change of $\bm{r}_p$ expressed in body-frame can be written as:
\begin{equation*}
    \dot{\bm{r}}_p = \bm{v}_p - \bm{v} - \S(\omega)\bm{r}_p
\end{equation*}
where $\vp \in \R^3$ is the linear velocity of the moving mass and $\bm{S}(\cdot) \in \R^{3 \times 3}$ is the skew-symmetric matrix. 
The total velocity of the system is then given by $\nu' = [\nu^\top, \bm{v}_p^\top]^\top$. 
This will be useful when describing the Newton-Euler equations of motion in section \ref{sec:kinetics}.

\subsection{Kinetics}
\label{sec:kinetics}
As described in \cite{Fos:21}, the kinetics of underwater vehicles is given by:
\begin{equation}
    \M\nuDot + \C(\nu)\nu + \D(\nu)\nu + \g(\eta) = \tau
    \label{eq:cams25-fossen}
\end{equation}
where $\M \in \R^{6 \times 6}$ is the mass matrix, $\C(\nu) \in \R^{6 \times 6}$ is the Coriolis matrix, $\D(\nu) \in \R^{6 \times 6}$ is the linear damping matrix, $\g(\eta) \in \R^6$ is the hydrostatic forces and $\tau \in \R^6$ is the vector containing external forces and torques.

In the literature, the mass and Coriolis terms are typically divided into rigid-body and added mass terms, i.e. $\M = \M_{RB} + \M_{A}$ and $\C(\nu) = \C_{RB}(\nu) + \C_{A}(\nu)$.
For a system with a moving mass actuator $m_p$ located at the point $\bm{r}_p$, the rigid-body mass can be expressed as:
{\small
    \begin{equation*}
        \begin{split}
        \M_{RB}\left(\bm{r}_p\right) = \\
        \begin{bmatrix}
            m\mathbb{I}_3 & - m_{s}\bm{S}(\bm{r}_{s}) -m_p\bm{S}(\bm{r}_p) & m_p \mathbb{I}_3\\
            m_{s}\bm{S}(\bm{r}_{s}) + m_p\bm{S}(\bm{r}_p)  & \bm{I}_b - m_p\bm{S}^2(\bm{r}_p)& m_p\bm{S}(\bm{r}_p)\\
            m_p \mathbb{I}_3 & -m_p\bm{S}(\bm{r}_p) & m_p \mathbb{I}_3
        \end{bmatrix}
        \end{split}
    \end{equation*}
}where $\mathbb{I}_3$ denotes the 3-by-3 identity matrix and $\bm{I}_b = \bm{I}_g - m_{s}\bm{S}^2(\bm{r}_{s})$ is the vessel inertia of the fixed body mass shifted to the CO. 
This mass matrix is the same as given in \cite{Woo:02}, but differs in the contribution of the static mass $m_s$. 
In \cite{Woo:02}, the static mass is said to be located at the CG, i.e. $\bm{r}_s = \bm{r}_g$, implying that moving $m_p$ impacts the position of $m_s$. 
This is a reasonable assumption if moving $m_p$ results in a negligible change in $\bm{r}_g$.
Yet, in this paper the location of the stationary mass is denoted separately and does not necessarily coincide with $\bm{r}_g$.

We choose to split the rigid body mass into a stationary matrix $\M_S$ and a changing matrix $\M_P(\bm{r}_p)$, given by \eqref{eq:cams25-static_mass_matrix} and \eqref{eq:cams25-moving_mass_matrix}, respectively.
\begin{equation}
    \label{eq:cams25-static_mass_matrix}
    \M_{S} =
    \begin{bmatrix}
        m\mathbb{I}_3 & - m_{s}\bm{S}(\bm{r}_{s}) & m_p \mathbb{I}_3\\
        m_{s}\bm{S}(\bm{r}_{s}) & \bm{I}_b & \bm{0}_{3 \times 3}\\
        m_p\mathbb{I}_3 & \bm{0}_{3 \times 3} & m_p \mathbb{I}_3
    \end{bmatrix}
\end{equation}
\begin{equation}
    \label{eq:cams25-moving_mass_matrix}
    \M_{P}(\bm{r}_p) =
    \begin{bmatrix}
        \bm{0}_{3 \times 3} & -m_p\bm{S}(\bm{r}_p) & \bm{0}_{3 \times 3}\\
        m_p\bm{S}(\bm{r}_p)  & - m_p\bm{S}^2(\bm{r}_p)& m_p\bm{S}(\bm{r}_p)\\
        \bm{0}_{3 \times 3} & -m_p\bm{S}(\bm{r}_p) & \bm{0}_{3 \times 3}
    \end{bmatrix}
\end{equation}

As mentioned in section \ref{sec:cams25-kinematics}, we assume the moving mass actuator is within the enclosed space of the vessel and therefore does not contribute to the added mass. 
Thus, the added mass matrix can be written as:
\begin{equation*}
    \M_A =
    \begin{bmatrix}
        \bm{A}_{11} & \bm{A}_{12} & \bm{0}_{3 \times 3} \\
        \bm{A}_{21} & \bm{A}_{22} & \bm{0}_{3 \times 3} \\
        \bm{0}_{3 \times 3} & \bm{0}_{3 \times 3} & \bm{0}_{3 \times 3}
    \end{bmatrix}
\end{equation*}
where $\bm{A}_{ij} \in \R^{3 \times 3}, i, j = \left\{ 1, 2 \right\}$ are sub-matrices. 
We can further assume $xy$-plane symmetry, which yields the convenient property of symmetry in added mass, the matrix is also assumed to be positive semi-definite, i.e. $\M_A = \M_A^\top \geq 0$. 
In an attempt to stay consistent with the literature, we choose to denote the total \emph{stationary} mass as $\M = \M_s + \M_A$ and the total mass including the internal moving mass $\M' = \M + \M_P$. The movable mass $m_p$ is assumed to be significantly smaller than the stationary mass $m_s$ and the movement $\bm{\dot{r}}_p$ is assumed to be slow, i.e. $m_p << m_s$ and $\bm{\dot{r}}_p < \varepsilon$. From this follows $\bm{\dot{M}}' = 0$.

The Coriolis-centripital forces of \eqref{eq:cams25-fossen} were derived in \cite{Sag:91} without an internal moving mass actuator.
For completeness, we will go through the calculation including a moving point mass here. 
In order to derive the Coriolis-centripital forces of the total system, we first look at the total kinetic energy of the system:
\begin{equation}
    \label{eq:kinetic-energy}
    \bm{T} = \frac{1}{2}\nu'^\top\M'\nu'
\end{equation}
Dividing $\nu'$ into its respective sub-vectors yields:
\begin{equation*}
    \begin{split}
        \bm{T} & = \frac{1}{2} \bm{v}^\top\M'_{11}\bm{v} + \frac{1}{2} \bm{v}^\top\M'_{12}\bm{\omega} + \frac{1}{2} \bm{v}^\top\M'_{13}\bm{v}_p \\
        & + \frac{1}{2} \bm{\omega}^\top\M'_{21}\bm{v} + \frac{1}{2} \bm{\omega}^\top\M'_{22}\bm{\omega} + \frac{1}{2} \bm{\omega}^\top\M'_{23}\bm{v}_p \\ 
        & + \frac{1}{2} \bm{v}_p^\top\M'_{31}\bm{v} + \frac{1}{2} \bm{v}_p^\top\M'_{32}\bm{\omega} + \frac{1}{2} \bm{v}_p^\top\M'_{33}\bm{v}_p
    \end{split}
\end{equation*}
Due to the symmetry of $\bm{M}'$, the partial derivatives of \eqref{eq:kinetic-energy} become:
{\small
\begin{align*}
    \begin{split}
        \frac{\partial \bm{T}}{\partial \bv} & = \M'_{11}\bv + \M'_{12}\omega + \M'_{13}\vp\\
        & = \left( m\mathbb{I}_3  + \bm{A}_{11}\right)\bv \\
        & \quad + \left( \bm{A}_{12} - m_s\bm{S}(\bm{r}_s) - m_p\bm{S}(\bm{r}_p) \right)\omega + m_p\mathbb{I}_3\vp
    \end{split}\\
    \begin{split}
        \frac{\partial \bm{T}}{\partial \omega} & = \M'_{21}\bv + \M'_{22}\omega + \M'_{23}\vp\\
        & = \left( m_s\bm{S}(\bm{r}_s) + m_p\bm{S}(\bm{r}_p) + \bm{A}_{21}\right)\bv \\ 
        & \quad + \left( \bm{I}_b - m_p\bm{S}(\bm{r}_p) + \bm{A}_{22} \right)\omega - m_p\bm{S}(\bm{r}_p)\vp
    \end{split}\\
    \begin{split}
        \frac{\partial \bm{T}}{\partial \vp} &= \M'_{31}\bv + \M'_{32}\omega + \M'_{33}\vp\\
        & = m_p\mathbb{I}_3\bv + m_p\bm{S}(\bm{r}_p)\omega + m_p\mathbb{I}_3\vp
    \end{split}
\end{align*}
}Using Kirchhoff as written by \cite{Lam:32} we get:
\begin{align*}
    \frac{d}{dt}\frac{\partial \bm{T}}{\partial \bv} + \bm{S}(\omega)\frac{\partial \bm{T}}{\partial \bv} & = \tau_{\bv} \\
    \frac{d}{dt}\frac{\partial \bm{T}}{\partial \omega} + \bm{S}(\bv)\frac{\partial \bm{T}}{\partial \bv} + \bm{S}(\omega)\frac{\partial \bm{T}}{\partial \omega} + \bm{S}(\bv_p)\frac{\partial \bm{T}}{\partial \bv_p} & = \tau_{\omega} \label{eq:second-term} \\
    \frac{d}{dt}\frac{\partial \bm{T}}{\partial \vp} + \bm{S}(\omega)\frac{\partial \bm{T}}{\partial \vp} & = \tau_{\vp}
\end{align*}
From this emerges a model on the form:
\begin{equation}
    \label{eq:cams25-temp_model}
    \M'\nuDot' + \C(\nu')\nu' = \tau'
\end{equation}
where $\tau' = [ \tau_{\bv}^\top, \tau_{\omega}^\top, \tau_{\bv_p}^\top ]^\top \in \R^9$ is the control input vector. Using the fact that $\bm{S}(\bm{a})\bm{b} = -\bm{S}(\bm{b})\bm{a}$, the Coriolis-centripital forces are given by:
{\small
\begin{equation}
    \label{eq:cams25-Coriolis}
    \C'(\nu')\nu' = 
    \begin{bmatrix}
        \bm{0}_{3 \times 3} && 
        -\bm{S}\left(\frac{\partial \bm{T}}{\partial \bv}\right) && 
        \bm{0}_{3 \times 3} \\
        -\bm{S}(\frac{\partial \bm{T}}{\partial \bv}) && 
        -\bm{S}\left(\frac{\partial \bm{T}}{\partial \omega}\right) && 
        -\bm{S}\left(\frac{\partial \bm{T}}{\partial \bv_p}\right)\\
        \bm{0}_{3 \times 3} && 
        -\bm{S}\left(\frac{\partial \bm{T}}{\partial \bv_p}\right) && 
        \bm{0}_{3 \times 3}
    \end{bmatrix}
    \begin{bmatrix}
        \bv\\
        \omega\\
        \vp
    \end{bmatrix}
\end{equation}
}Like with the mass matrix expression, \eqref{eq:cams25-Coriolis} can be restructured into a static and moving terms, we denote this $\C'(\nu') = \C(\nu') + \C_P(\nu', \bm{r}_p)$, where $\C(\nu')$ term already contains the Coriolis-centripital forces exerted by added mass. 
As proved by \cite{Sag:91}, the Coriolis parametrization $\C'(\nu')$ is not unique.
An example of the parametrization of $\C(\nu')$ and $\C_P(\nu', \bm{r}_p)$ can be found in \eqref{eq:cams25-C} and \eqref{eq:cams25-C_P}, respectively.

\subsection{Hydrostatics}
Forces on the vehicle due to gravity are not captured by using Kirchhoff's equations. 
Moving the CG has a great impact on the hydrostatic behaviour of underwater vehicles. 
The gravitational forces acting on the stationary and moving mass are given by:
\begin{equation*}
    W_s = m_sg, \quad W_p = m_pg
\end{equation*}
where $g$ is the acceleration due to gravity. 
The buoyancy of the vessel, given by the water density $\rho$ and the volume of total water displaced $\nabla$, is expressed as:
\begin{equation*}
    B = \rho g \nabla
\end{equation*}
These two forces act in opposite directions on the CO and CB, respectively. 
The linear forces, in the inertial frame, for the vessel with total mass $m$ are therefore given by:
\begin{equation*}
    \bm{f}^n = \bm{f}_s^n + \bm{f}_p^n - \bm{f}_b^n
\end{equation*}
where $\bm{f}_s^n = \left[ 0, 0, W_s \right]^\top$, $\bm{f}_p^n = [ 0, 0, W_p ]^\top$ and $\bm{f}_b^n = \left[ 0, 0, B \right]^\top$, respectively. 
Rotating this to the body-fixed frame yields:
\begin{equation*}
    \begin{split}
        \bm{f} & = \bm{R}(\bm{\Theta})^\top\bm{f}_s^n + \bm{R}(\bm{\Theta})^\top\bm{f}_p^n - \bm{R}(\bm{\Theta})^\top\bm{f}_b^n \\
        & = \bm{f}_s + \bm{f}_p - \bm{f}_b
    \end{split}
\end{equation*}

Induced by these linear forces are the torques around the CO.
Combining these results in a single vector yields:
\begin{equation*}
    \bm{g}'(\eta, \bm{r}_p) = -
    \begin{bmatrix}
        \bm{f}_s + \bm{f}_p - \bm{f}_b \\
        \bm{S}(\bm{r}_s) \bm{f}_s + \bm{S}(\bm{r}_p) \bm{f}_p + \bm{S}(\bm{r}_b)(-\bm{f}_b) \\
        \bm{f}_{p}
    \end{bmatrix}
\end{equation*}
Moreover, since it is assumed that CB = CO and that the system is neutrally buoyant, the resulting hydrostatic expression becomes:
\begin{equation}
    \label{eq:cams25-hydrostatics}
    \bm{g}'(\eta, \bm{r}_p) = -
    \begin{bmatrix}
        \bm{0}_3 \\
        \bm{S}(\bm{r}_s) \bm{R}(\bm{\Theta})^\top\bm{f}_s^n + \bm{S}(\bm{r}_p) \bm{R}(\bm{\Theta})^\top\bm{f}_p^n \\
        \bm{R}(\bm{\Theta})^\top\bm{f}_p^n
    \end{bmatrix}
\end{equation}
Note the negative sign of \eqref{eq:cams25-hydrostatics}, this allows the restoring forces to appear on the left side of \eqref{eq:cams25-temp_model}.
The hydrostatics can also be separated into fixed and moving mass terms:
\begin{equation}
    \label{eq:cams25-g_s}
    \bm{g}(\eta) = -
    \begin{bmatrix}
        \bm{0}_3 \\
        \bm{S}(\bm{r}_s) \bm{R}(\bm{\Theta})^\top\bm{f}_s^n \\
        \bm{0}_3
    \end{bmatrix}
\end{equation}
\begin{equation}
    \label{eq:cams25-g_p}
    \bm{g}_P(\eta, \bm{r}_p) = -
    \begin{bmatrix}
        \bm{0}_3 \\
        \bm{S}(\bm{r}_p) \bm{R}(\bm{\Theta})^\top\bm{f}_p^n \\
        \bm{R}(\bm{\Theta})^\top\bm{f}_p^n
    \end{bmatrix}
\end{equation}
The final Newton-Euler equations of motion then becomes:
\begin{equation}
    \label{eq:cams25-generic-model}
    \begin{split}
        \etaDot & = \J(\eta)\nu \\
        \dot{\bm{r}}_p & = \bm{v}_p - \bm{v} - \S(\omega)\bm{r}_p \\
        \left(\M + \M_P(\bm{r}_p)\right)\nuDot' & + \left(\C(\nu') + \C_P(\nu', \bm{r}_p)\right)\nu'\\  
        & + \g(\eta) + \bm{g}_P(\eta, \bm{r}_p) = \tau'
    \end{split}
\end{equation}
Drag and damping forces are not affected by an internal moving mass actuator and are vehicle specific. For reasons of generality and clarity of the model, we have chosen to omit these terms.

\section{Practical Example}
To validate our proposed model \eqref{eq:cams25-generic-model}, we will give a practical example applying it to the Remus 100 AUV and comparing this in simulation to the Hamiltonian formulation in \cite{Woo:02}. The resulting simulation code can be found in \cite{Cams25Ram:25}.
\subsection{Remus Model with Moving Mass Actuator}
\label{sec:cams25-remus}
Values for simulations are those given in \cite{All:00} and \cite{FosPy:25}. 
For this example we constrain the moving mass $m_p$ to only move along the centreline, $5$cm below the CO and moving $\pm5$cm back and forth, i.e.  $\bm{r}_p = [x_p, 0, 0.05]^\top$ and $ x_p \in \left[-0.05, 0.05\right]$ (denoted in metres).
The stationary mass $m_s$ is assumed to be located at the CO, i.e. $\bm{r}_s = \bm{0}_{3}$. The magnitude of the moving mass is set to one sixth of the total mass, i.e. $m_s = \frac{1}{6}m$.

\begin{figure}[t]
    \centering
    \includegraphics[width=0.48\textwidth]{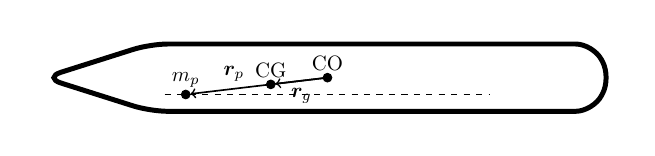}
    \caption{The Remus 100 vehicle with moving mass $m_p$ moving along the dotted line.
    Since $m_s$ is located at the CO, the CG will always be directly between the CO and $m_p$.}
    \label{fig:cams25-remus}
\end{figure}

As illustrated in \figref{fig:cams25-remus}, it follows that the CG is always located directly between the origin and the moving mass.
We assume the vehicle is neutrally buoyant and approximate the hull to be an ellipsoid with an evenly distributed static mass $m_s$.
The inertia of the rigid body around the CO is given by:
\begin{equation*}
    \bm{I}_g = \text{diag}\left\{ \frac{2}{5}m_sb^2, \frac{1}{5}m_s(a^2 + b^2), \frac{1}{5}m_s(a^2 + b^2) \right\}
\end{equation*}
where $a \in \R$ represents the forward semi-axis and $b \in \R$ represent the two remaining semi-axes of the ellipsoid (\cite{Fos:21}). 
From the ellipsoid symmetries it also follows that the added mass term simplifies to:
\begin{equation*}
    \M_A =
    \begin{bmatrix}
        \bm{A}_{1} & \bm{0}_{3 \times 3} & \bm{0}_{3 \times 3} \\
        \bm{0}_{3 \times 3} & \bm{A}_{2} & \bm{0}_{3 \times 3} \\
        \bm{0}_{3 \times 3} & \bm{0}_{3 \times 3} & \bm{0}_{3 \times 3}
    \end{bmatrix}
\end{equation*}
where the sub-matrices $\bm{A}_{1} = -\text{diag}\left\{ X_{\dot{u}}, Y_{\dot{v}}, Z_{\dot{w}} \right\}$ and $\bm{A}_{2} = -\text{diag}\left\{ K_{\dot{u}}, M_{\dot{v}}, N_{\dot{w}} \right\}$ are the added mass of linear and angular motion, respectively. Under the presented assumptions the mass matrix $\M$ for the Remus 100 AUV is: 
\begin{equation}
    \label{eq:cams25-remus-mass_matrix}
    \M =
    \begin{bmatrix}
        m\mathbb{I}_3 + \bm{A}_1 & \bm{0}_{3 \times 3} & m_p\mathbb{I}_{3}\\
        \bm{0}_{3 \times 3} & \bm{I}_g + \bm{A}_2 & \bm{0}_{3 \times 3}\\
        m_p\mathbb{I}_3 & \bm{0}_{3 \times 3} & m_p\mathbb{I}_3
    \end{bmatrix}
\end{equation} 
With the moving mass matrix $\M_P$ still given by \eqref{eq:cams25-moving_mass_matrix}.
The Coriolis matrices, $\C(\nu')$ and $\C_P(\nu', \bm{r}_p)$, for the Remus 100 AUV are given by \eqref{eq:cams25-C_remus} and \eqref{eq:cams25-C_P}, respectively.

\subsection{Hydrostatic Considerations}
\label{sec:cams25-hydrostatic-considerations}
For ease of simulating the total system, we add a constraining control force directly counteracting the force of gravity pulling on $m_p$. 
Since the rigid body mass is located at the CO the hydrostatics term $\bm{g}'(\eta, \bm{r}_p)$ for the Remus AUV simplifies to:
\begin{equation}
    \bm{g}(\bm{r}_p) = -
    \begin{bmatrix}
        \bm{0}_3 \\
        \bm{S}(\bm{r}_p) \bm{R}(\bm{\Theta})^\top\bm{f}_p^n \\
        \bm{0}_3
    \end{bmatrix}
\end{equation}

\section{Simulation}
In the following section, the model presented in this paper is compared to that of \cite{Woo:02}.
As an open loop test, external forces are applied to the vessel in surge direction and to the internal moving mass.
The formulation by \cite{Woo:02} has been altered to include these external forces as well as including the hydrostatic considerations mentioned in \ref{sec:cams25-hydrostatic-considerations}.

The test is conducted by a $1/2$ Newton force being applied to $m_p$ in the forward direction, denoted $\scalarTau_{X_p}$, when the vehicle reaches at depth of $20$ metres, a $-1/2$ Newton force is applied instead. When the vehicle reaches $3$ metres depth, the direction of $\scalarTau_{X_p}$ is switched again. This pattern is repeated for 500 seconds.
A $1$ Newton surge force, denoted $\scalarTau_{X}$, is applied to the vehicle throughout the simulations.
\subsection{Results}
\begin{figure}[t]
    \centering
    \includegraphics[width=0.487\textwidth]{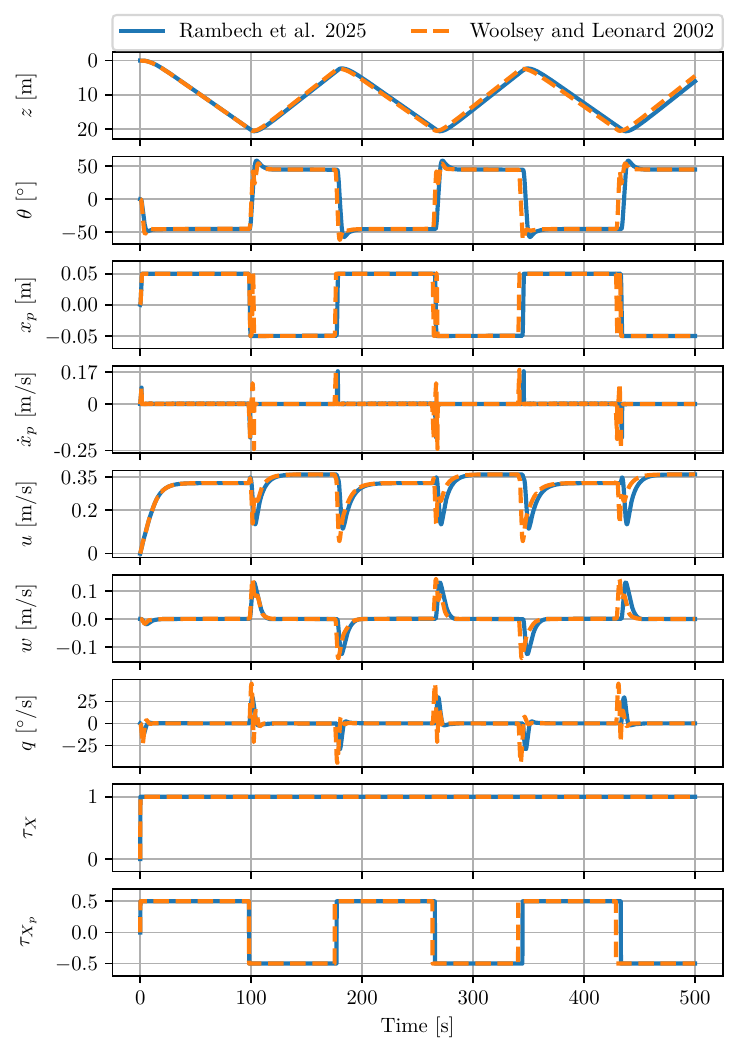}
    \caption{Simulation results of the proposed model compared to that of \cite{Woo:02}.}
    \label{fig:cams25-simulations}
\end{figure}
\begin{figure}[t]
    \centering
    \includegraphics[width=0.487\textwidth]{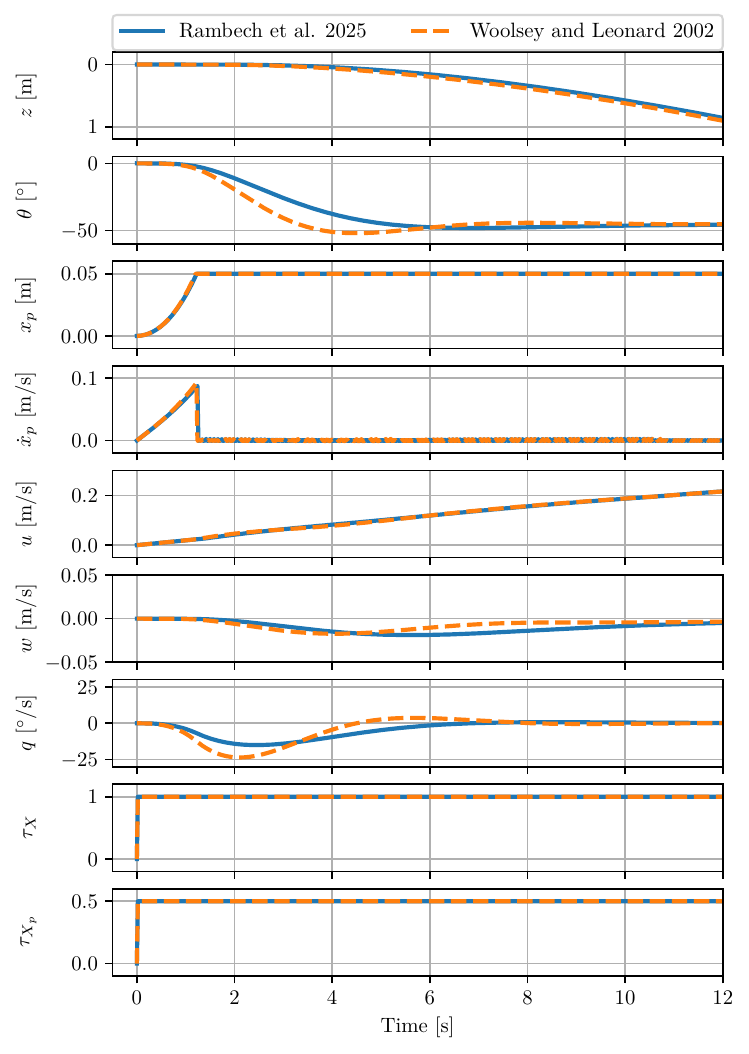}
    \caption{Simulation results of the proposed model compared to that of \cite{Woo:02}. Close-up of the first 12 seconds of simulation. Initial conditions are zero.}
    \label{fig:cams25-simulations-closeup}
\end{figure}
The results from the simulations are shown in \figref{fig:cams25-simulations} with a close up of initial response shown in \figref{fig:cams25-simulations-closeup}. 
We observe that moving $m_p$ impacts the pitch of the vehicle as expected.
From \figref{fig:cams25-simulations}, the surge speed $u$ is higher for both formulations when the vehicle is pitched upwards.
This is due to the coupling in forces, where a negative $\scalarTau_{X_p}$ yields a positive contribution in surge and vice versa.

The close-up in \figref{fig:cams25-simulations-closeup} clearly shows that the pitch velocity $q$ of the AUV is higher for the \cite{Woo:02} formulation. 
The difference between the two models can be attributed to the calculation of the torque produced around the CO. 
In our proposed formulation, the torque around CO changes less rapidly due to $\bm{r}_s$, which is zero in this example, being used as a lever arm for the static mass $m_s$ as opposed to $\bm{r}_g$.
If $\bm{r}_g$ is replaced by $\bm{r}_s$ in the \cite{Woo:02} formulation, the two models become equivalent. 
Our proposed model is thus validated with existing literature.

\section{Conclusion}
In this paper we have established a generalized vectorial Newton-Euler formulation of underwater vehicles with internal moving mass actuators. 
The model builds on earlier manoeuvring models and enable moving mass actuators to be expressed using modern vectorial notation. The model has also been validated through open loop simulations and compared to the Hamiltonian formulation given in \cite{Woo:02}.

Further research may be conducted to demonstrate the practicality of the model by developing controllers exploiting the symmetric and skew-symmetric properties of \eqref{eq:cams25-moving_mass_matrix} and \eqref{eq:cams25-Coriolis}, respectively.

\bibliography{ifacconf}

\appendix
\section{Coriolis matrices}
{\small
\begin{equation}
    \label{eq:cams25-C}
    \begin{split}
    \C(\nu') = \left[\begin{matrix}
            \bm{0}_{3 \times 3} \\
            -m\S(\bv) - \S(\bm{A}_{11}\bv) - m_s\S(\S(\omega)r_s)\\
            \bm{0}_{3 \times 3}
        \end{matrix} \right. \qquad \qquad \qquad \\
        \quad \left.\begin{matrix}
            -m\S(\bv) - \S(\bm{A}_{11}\bv) - m_s\S(\S(\omega)r_s) & 
            \bm{0}_{3 \times 3}\\
            -m_s\S(\S(\bm{r}_s)\bv) - \S(\bm{A}_{21}\bv) - \S(\bm{I}_b\omega) - \S(\bm{A}_{22}\omega) & 
            -m_p\S(\bv)\\
            -m_p\S(\bv) & 
            \bm{0}_{3 \times 3}
        \end{matrix}\right]
    \end{split}
\end{equation}
}
{\small
\begin{equation}
    \label{eq:cams25-C_P}
    \begin{split}
    \C_P(\nu', \bm{r}_p) = \left[\begin{matrix}
            \bm{0}_{3 \times 3} \\
            -m_p\S(\S(\omega)\bm{r}_p + \vp) \\
            \bm{0}_{3 \times 3}
        \end{matrix} \right.\qquad \qquad \qquad \qquad \qquad \\
        \left.\begin{matrix}
            -m_p\S(\S(\omega)\bm{r}_p + \vp) & 
            \bm{0}_{3 \times 3}\\
            -m_p\S(\S(\bm{r}_p)\bv + \S(\omega)\bm{r}_p + \S(\vp)\bm{r}_p) & 
            -m_p\S(\S(\bm{r}_p)\omega + \vp) \\
            -m_p\S(\S(\bm{r}_p)\omega + \vp) & 
            \bm{0}_{3 \times 3}
        \end{matrix}\right]
    \end{split}
\end{equation}
}  
{\small
\begin{equation}
    \label{eq:cams25-C_remus}
    \begin{split}
    \C(\nu') &= \left[\begin{matrix}
            \bm{0}_{3 \times 3} \\
            -m\S(\bv) - \S(\bm{A}_{1}\bv) - m_s\S(\S(\omega)r_s) \\
            \bm{0}_{3 \times 3}
        \end{matrix} \right. \\
        & \left.\begin{matrix}
            -m\S(\bv) - \S(\bm{A}_{1}\bv) - m_s\S(\S(\omega)r_s) & 
            \bm{0}_{3 \times 3}\\
            -m_s\S(\S(\bm{r}_s)\bv) - \S(\bm{I}_b\omega) - \S(\bm{A}_{2}\omega) & 
            -m_p\S(\bv)\\
            -m_p\S(\bv) & 
            \bm{0}_{3 \times 3}
        \end{matrix}\right]
    \end{split}
\end{equation}
}
\end{document}